\documentclass[runningheads]{llncs}

\usepackage[mobile]{eccv}

\usepackage{eccvabbrv}

\usepackage{graphicx}
\usepackage{booktabs}

\usepackage[accsupp]{axessibility}  %

\usepackage{hyperref}

\usepackage{orcidlink}

\usepackage{amsmath}
\usepackage{amssymb}
\usepackage{wrapfig}
\usepackage{subcaption}
\usepackage{colortbl}
\definecolor{mygray}{gray}{.92}

\newcommand{\tabref}[1]{Table \ref{#1}}

\newcommand{\figref}[1]{Fig. \ref{#1}}

\newcommand{\myPara}[1]{~\textbf{#1}}

\def\ie{\emph{i.e.}}
\def\eg{\emph{e.g.}}

\def\etal{{\em et al.~}}

\def\ourmodel{CGRSeg}

\usepackage[misc]{ifsym}
\begin{document}
\title{Context-Guided Spatial Feature Reconstruction for Efficient Semantic Segmentation} 

\titlerunning{Context-Guided Spatial Feature Reconstruction for Efficient Segmentation}

\renewcommand\footnotemark{}
\author{Zhenliang Ni\orcidlink{0000-0002-3358-1994} \and
Xinghao Chen$^{(\textrm{\Letter})}$\orcidlink{0000-0002-2102-8235} \and  
Yingjie Zhai\orcidlink{0000-0001-7593-6055} \and 
Yehui Tang\orcidlink{0000-0002-0322-4283}  \and \\
Yunhe Wang$^{(\textrm{\Letter})}$\orcidlink{0000-0002-0142-509X}
\thanks{$\textrm{\Letter}$~Corresponding authors.}
}

\authorrunning{Z. Ni et al.}

\institute{Huawei Noah’s Ark Lab\\
\email{\{xinghao.chen, yunhe.wang\}@huawei.com}}

\maketitle

\begin{abstract}
  Semantic segmentation is an important task for numerous applications but it is still quite challenging to achieve advanced performance with limited computational costs. In this paper, we present CGRSeg, an efficient yet competitive segmentation framework based on context-guided spatial feature reconstruction. 
  A Rectangular Self-Calibration Module is carefully designed for spatial feature reconstruction and pyramid context extraction. It captures the axial global context in both horizontal and vertical directions to explicitly model rectangular key areas. A shape self-calibration function is designed to make the key areas closer to foreground objects. 
  Besides, a lightweight Dynamic Prototype Guided head is proposed to improve the classification of foreground objects by explicit class embedding.
  Our CGRSeg is extensively evaluated on ADE20K, COCO-Stuff, and Pascal Context benchmarks, and achieves state-of-the-art semantic performance. Specifically, it achieves $43.6\%$ mIoU on ADE20K with only $4.0$ GFLOPs, which is $0.9\%$ and $2.5\%$ mIoU better than SeaFormer and SegNeXt but with about $38.0\%$ fewer GFLOPs. Code is available at \url{https://github.com/nizhenliang/CGRSeg}.
\end{abstract}

\section{Introduction}
Semantic segmentation is a foundational computer vision task that aims to assign semantic labels to each pixel in an image. In recent years, deep learning methods have achieved remarkable results in semantic segmentation, thanks to the availability of large-scale datasets, powerful computing resources, and advanced network architectures. Early studies for semantic segmentation are mainly based on convolution neural networks (CNNs), such as FCN~\cite{fcn}, PSPNet~\cite{pspnet}, and DeepLab~\cite{deeplab,deeplabv2,deeplabv3,autodeeplab}. More recently, a series of methods for semantic segmentation have been proposed based on vision transformer and achieve competitive performance~\cite{setr,segformer,segvit,seaformer}. Deep learning methods have significantly improved the accuracy of semantic segmentation. However, it is still quite a challenging topic for how to achieve advanced segmentation performance with limited computing resources.

\begin{figure}[tb]
	\begin{minipage}[t]{.5\textwidth}
		\centering
		\includegraphics[width=\textwidth]{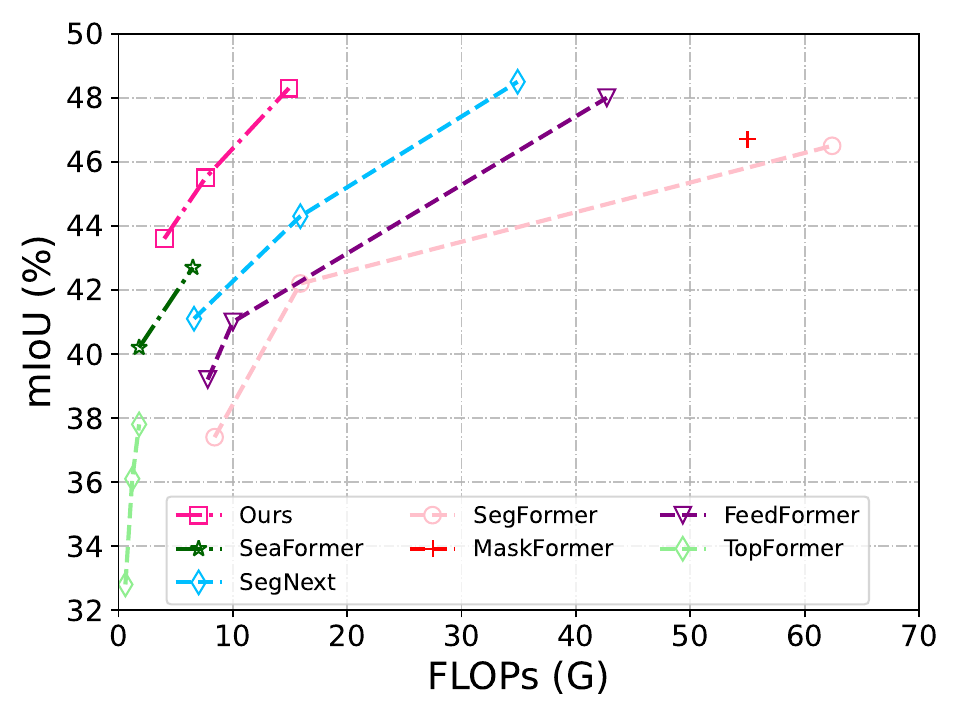}
	\end{minipage}
	\begin{minipage}[t]{.5\textwidth}
		\centering
		\includegraphics[width=\textwidth]{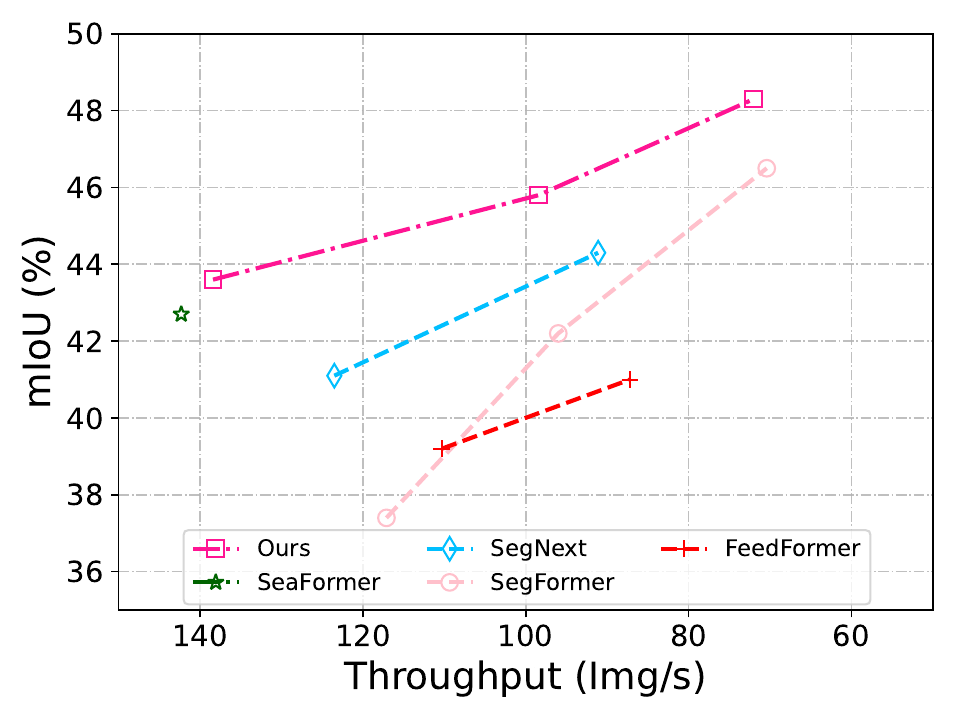}
	\end{minipage}  
	\caption{\textbf{Performance \textit{vs.} FLOPs and Throughput on ADE20K~\cite{ade20k}}. Our model achieves better trade-off between accuracy and computational cost than prior methods. Moreover, our model outperforms other models in throughput with higher accuracy.}
	\label{iou_flops}
\end{figure}

Efficient and lightweight semantic segmentation is an important topic and has attracted great research interest. 
Yu~\etal\cite{bisenet} introduced a two-pathway architecture named BiSeNet, which captures both local and global information and achieves a good balance between speed and performance. 
SegFormer~\cite{segformer} designed a hierarchical Transformer encoder coupled with a lightweight multilayer perception (MLP) decoder to achieve both high precision and speeds. 
Wan~\etal~\cite{seaformer} introduced an efficient squeeze-enhanced axial transformer (SeaFormer) to reduce the computational complexity of the standard transformer. Recently, 
Guo~\etal~\cite{segnext} proposed SegNeXt to demonstrate that an encoder with simple and cheap convolutions can still achieve competitive performance.

Nevertheless, existing lightweight models, constrained by their limited feature representation capabilities, often encounter difficulties in modeling the boundaries and distinguishing the categories of foreground objects. These limitations lead to inaccuracies in boundary segmentation and misclassification.
To address these problems, we carefully design the Rectangular Self-Calibration Module (RCM) to improve the position modeling of foreground objects. Besides, a Dynamic Prototype Guided (DPG) head is introduced to embed class information to enhance the class discrimination of foreground objects. Moreover, to further introduce the pyramid context to improve the feature representation, we design a context-guided spatial feature reconstruction network (CGRSeg) consisting of pyramid context extraction, spatial feature reconstruction, and a lightweight head. 

The Rectangular Self-Calibration Module is proposed to improve the ability to locate foreground objects and extract pyramid context. The rectangular self-calibration attention is a core component of RCM. It adopts horizontal pooling and vertical pooling to capture the axial global context and generate two axis vectors. The two axis vectors are added together to model a rectangular attention region. A shape self-calibration function is designed to adjust the shape of rectangular attention and bring it closer to the foreground feature, which consists of large-kernel strip convolution. Besides, a fusion function is designed to fuse attention features and increase local detail. In this way, RCM can make the model focus more on the foreground for spatial feature reconstruction, while also effectively capturing the axial global context for pyramid context extraction.

The Dynamic Prototype Guided head is proposed to improve the classification of the foreground objects via explicit class embedding. Specifically, the features are first projected into the class feature space. Then, the feature in class space and the feature in pixel space are multiplied to obtain the dynamic prototype. The dynamic prototype can reflect the feature distribution of all classes on each image. To further embed class information, the dynamic prototype is compressed to obtain the class embedding vector. Finally, the class embedding vector is projected into the pixel feature space, and the pixel features are weighted to enhance the discrimination between different classes. In this way, the DPG head can effectively enhance class discrimination and improve classification performance.

The proposed \ourmodel~achieves state-of-the-art segmentation performance on ADE20K, COCO Stuff, and Pascal Context with fewer FLOPs and higher Throughput. The performance on ADE20K is shown in Fig.~\ref{iou_flops}. Specifically, our method achieves $43.6\%$ mIoU on ADE20K with only $4.0$ GFLOPs, which is $0.9\%$ and $2.5\%$ mIoU better than SeaFormer~\cite{segformer} and SegNeXt~\cite{segnext} but with about $38.0\%$ fewer GFLOPs.

\section{Related Work}
\textbf{Efficient Semantic Segmentation}
The lightweight semantic segmentation methods based on CNN include BiSeNet series~\cite{bisenet}, STDCNet~\cite{stdnc}, DFANet~\cite{dfanet}, SegNeXt~\cite{segnext} and so on.
STDCNet~\cite{stdnc} designed an STDC module to extract scalable receptive field and multi-scale information and used boundary supervision to supplement spatial details. 
SegNeXt~\cite{segnext} captured spatial attention via multi-scale convolutional features, achieving better performance than vision transformers. 
In recent years, several Transformer-based methods have emerged in this field~\cite{clustseg, seaformer, topformer, segformer}.
TopFormer~\cite{topformer} introduced a hybrid architecture of CNN and Transformer, which took tokens from various scales as input to produce scale-aware semantic features.
SeaFormer~\cite{seaformer} designed a squeeze-enhanced axial transformer in the encoder to improve feature representation and proposed a new two-path structure.
Besides, there are some methods to improve the segmentation performance through contrast learning or other training approaches, such as HSSN~\cite{hssn}, and ContrastiveSeg~\cite{contrastseg}.
Different from these methods, our CGRSeg employs pyramid context-guided spatial feature reconstruction to improve the ability to model foreground objects, which is more efficient.

\textbf{Efficient Context Extraction Modules}
In previous works, attention is often used for context extraction~\cite{coordatt,ge,gcnet, nonlocal}, such as coordinate attention and GatherExcite. Coordinate attention~\cite{coordatt} captured the axial global context through horizontal and vertical pooling, which also extracted coordinate relationships to enhance positional awareness. GatherExcite~\cite{ge} efficiently aggregated contextual information across large neighborhoods to refine the feature representation. The proposed RCM uses addition to model the critical region, and designs shape self-calibration to calibrate the rectangle attention, which can make the model pay more attention to the foreground features. These designs allow RCM to outperform existing attention. With the development of transformers, some convolutional networks~\cite{convnext, inceptionnext} adopted transformer-like structures to enhance feature representation. ConvNeXt~\cite{convnext} first proposed the structure of large kernel depth-wise convolution combined with BN and MLP, which can significantly improve the feature representation of traditional convolutional neural networks. InceptionNeXt~\cite{inceptionnext} proposed the MetaNeXt architecture and combined the Inception block with the MetaNeXt architecture, significantly improving the model performance. These works show that the MetaNeXt structure has powerful context extraction ability.

\textbf{Segmentation Head}
Segmentation heads are usually designed to improve the performance of segmentation networks. 
Some segmentation heads use multi-scale features to achieve advanced performance, such as PSPNet~\cite{pspnet}, DeepLab series~\cite{deeplab,deeplabv2,deeplabv3,autodeeplab}.
PSPNet~\cite{pspnet} designed the pyramid pooling module to capture multi-scale information.
Some segmentation heads, such as OCRHead~\cite{ocrhead} and DAHead~\cite{danet}, captured long-range dependencies to improve performance.
Hamburger Head~\cite{hamburgerhead} introduced a powerful global context module based on matrix decomposition, surpassing various attention modules on semantic segmentation.
There is also some work to improve the model performance by improving the classifier~\cite{protoseg}.
ProtoSeg~\cite{protoseg} proposed a new perspective on semantic segmentation by replacing the pixel-wise classifier with the prototype-based classifier.
However, existing segmentation heads are not suitable for efficient segmentation because they require large computational resources and have poorer performance.

\section{Method}
\subsection{CGRSeg}
\begin{figure}[t]
	\centering
	\includegraphics[width=\textwidth]{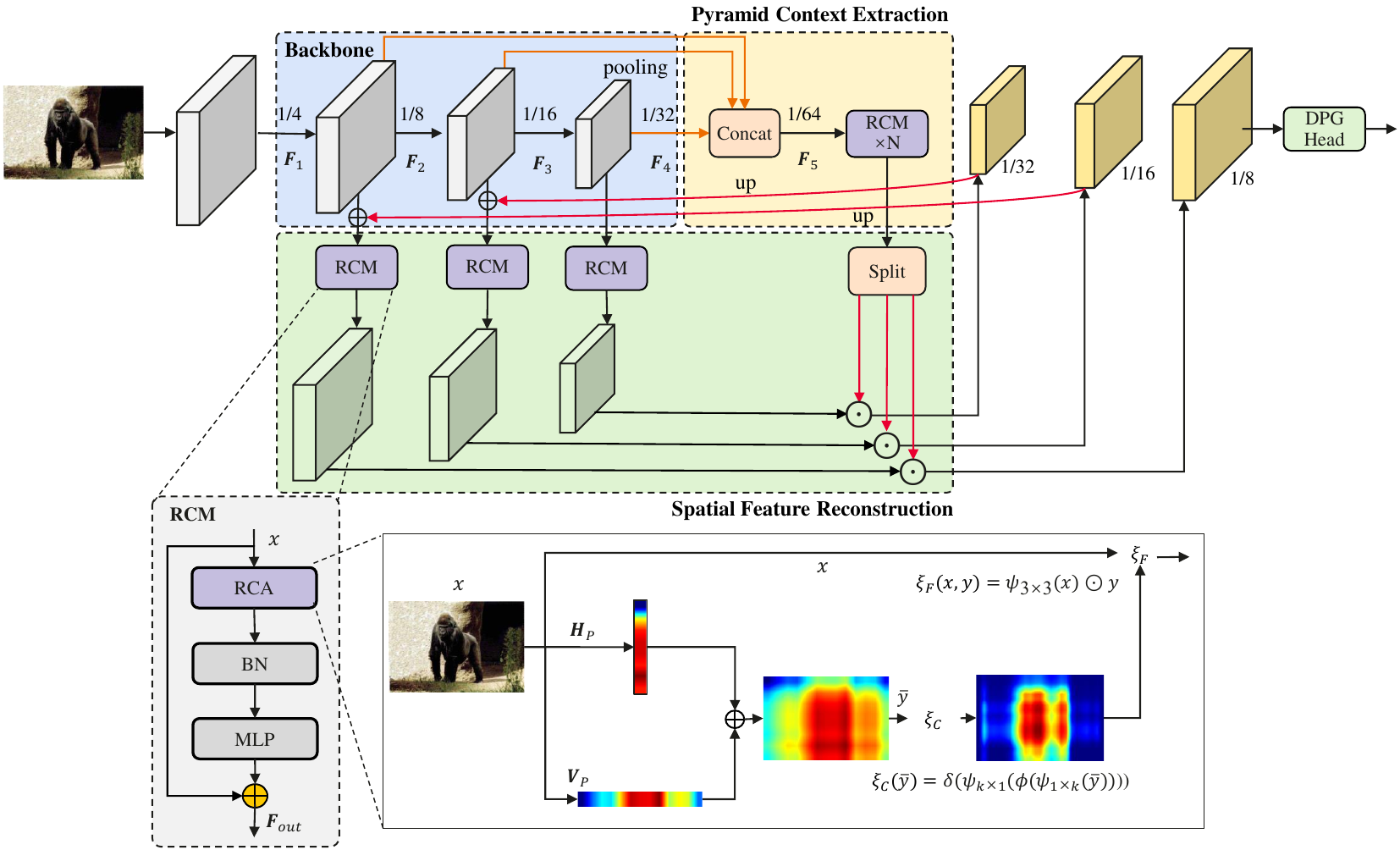}
	\caption{The overall architecture of \ourmodel. The Rectangular Self-Calibration Module (RCM) is designed for spatial feature reconstruction and pyramid context extraction. The rectangular self-calibration attention (RCA) explicitly models the rectangular region and calibrates the attention shape. The Dynamic Prototype Guided (DPG) head improves the classification of the foreground objects via explicit class embedding. }
	\label{fig:network}
\end{figure}

Previous work~\cite{topformer, seaformer, danet, deeplabv3plus} has shown that foreground object modeling and pyramid context extraction are critical for segmentation. To this end, we design a context-guided spatial feature reconstruction network (CGRSeg). As shown in \figref{fig:network}, CGRSeg consists of three key parts: pyramid context extraction, spatial feature reconstruction, and a lightweight head. To make the model focus on the foreground features, a Rectangular Self-Calibration Module (RCM) is proposed. Besides, it captures the axial global context for pyramid context extraction. A Dynamic Prototype Guided (DPG) head is proposed to improve the classification of the foreground objects via explicit class embedding. 

\textbf{Pyramid Context Extraction} The RCM is applied to extract pyramid context, which applies horizontal and vertical pooling to capture the axial context in two directions. Besides, RCM employs MLP to further enhance feature representation. As shown in \figref{fig:network}, a stepwise downsampling encoder is used in the proposed model. 
Encoder produces features of different scales features as $[\mathbf{F}_1$, $\mathbf{F}_2$, $\mathbf{F}_3$, $\mathbf{F}_4]$, with the resolution of $[\frac{H}{4}\times \frac{W}{4}$,$\frac{H}{8}\times \frac{W}{8}$,$\frac{H}{16}\times \frac{W}{16},\frac{H}{32}\times \frac{W}{32}]$, respectively. To ensure the efficiency of the whole network, the largest-scale feature, \ie, $\mathbf{F}_1$, is dropped by the decoder.
Then, the lower-scale features $\mathbf{F}_2$, $\mathbf{F}_3$ and $\mathbf{F_4}$ are down-sampled to $\frac{H}{64}\times \frac{W}{64}$ size by the average pooling, and they are concatenated together to generate the pyramid feature $\mathbf{F}_5$. $\mathbf{F}_5$ is fed into multiple stacked RCMs for pyramid feature interaction and extract scale-aware semantic features. Finally, the features are split and upsampled to the original scale after the pyramid feature extraction. 
This process can be formulated as:
\begin{equation}
\rm P= RCM(AP(\mathbf{F}_2,8),AP(\mathbf{F}_3,4),AP(\mathbf{F}_4,2)),
\end{equation}
where $AP(\mathbf{F}, x)$ represents the average pooling that down-samples the feature $\mathbf{F}$ by the factor $x$. $P$ is the feature with pyramid context.

\textbf{Spatial Feature Reconstruction} 
To make the decoder features more focused on the foreground, RCM is employed to reconstruct the spatial features. Specifically, the CGRSeg fuses the low-level spatial features from the encoder with the high-level features of the corresponding scale in the decoder. The fused features are reconstructed by using RCM. RCM captures the axial global context to model the rectangular critical region and then uses the shape self-calibration function to adjust the attention region to the foreground. Moreover, pyramid features are applied to guide the spatial feature reconstruction to make the reconstructed features feel multi-scale information.

\textbf{Dynamic Prototype Guided Head} 
To improve the classification of foreground objects, we introduce a dynamic prototype guided segmentation head. The segmentation head generates the dynamic prototype to embed class information. The class embedding helps to enhance the distinction between different classes and improve classification accuracy.

\subsection{Rectangular Self-Calibration Module}
The Rectangular Self-Calibration Module (RCM) is designed to make the model focus on the foreground. It can also capture the axial global context for pyramid context extraction. The module consists of rectangular self-calibration attention, batch normalization, and MLP.
\begin{figure}[t]
  \centering
  \includegraphics[width=0.7\columnwidth]{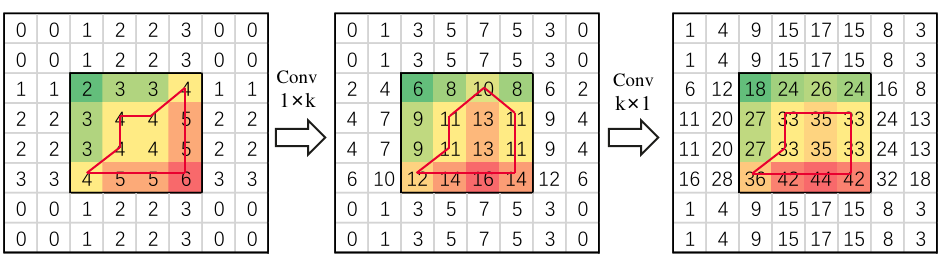}
  \caption{The shape change of the highlighted region is caused by the rectangular self-calibration attention. By optimizing the weights of the two strip convolutions during training, the attention region is calibrated closer to the foreground object.}
\label{strip_conv}
\end{figure}

The rectangular self-calibration attention (RCA) adopts horizontal pooling and vertical pooling to capture the axial global context in two directions, generating two distinct axis vectors. By applying broadcast addition to these two vectors, RCA can effectively model the rectangle region of interest.
Then, a shape self-calibration function is designed to calibrate the region of interest, which can make the region of interest closer to the foreground object. Here, two large-kernel strip convolutions are utilized to calibrate the attention map in the horizontal and vertical directions decoupledly. 
First, we employ the horizontal strip convolution to calibrate the shape in the horizontal direction, which adjusts each row of elements to make the horizontal shape closer to the foreground object. Then, the features are normalized by BN, and the non-linearity is added by ReLU. Subsequently, the shape is also calibrated by using the vertical strip convolution in the vertical direction. In this way, the convolution in two directions can be decoupled, which can adapt to any shape.

The weights of the strip convolutions are learnable. As shown in Fig.~\ref{strip_conv}, the shape of the highlighted area is changed by two strip convolutions with different weights. By training, this function can learn the appropriate weights to adjust the rectangular area to the foreground object. Furthermore, in Fig.~\ref{fig:network}, the visualization results show that the rectangular self-calibration attention does effectively model the rectangular key region and adjusts the features to focus more on the foreground by the shape self-calibration function.
The shape self-calibration function can be formulated as follows: 
\begin{equation}
  \xi_C(\overline{y}) =\delta (\psi_{k \times 1}(\phi (\psi_{1 \times k}(\overline{y} )))),
\end{equation}
where $\psi$ indicates the large-kernel strip convolution, and $k$ indicates the kernel size of the strip convolution. $\phi$ indicates the Batch Normalization followed by the ReLU function, and $\delta$ indicates the Sigmoid function.

Besides, a feature fusion function is designed to fuse attention features with input features. Specifically, 3 $\times$ 3 depth-wise convolution is used to further extract local details of the input feature. The calibrated attention feature is weighted to the refined input feature by Hadamard Product.
\begin{equation}
  \xi_F(x, y) =\psi_{3 \times 3}(x) \odot y ,
\end{equation}
where $\psi_{3 \times 3}$ indicates the deep-wise convolution with kernel $3 \times 3$. y is the attention feature obtained in the previous step. $\odot$ is the Hadamard Product.

Recently, ConvNeXt~\cite{convnext} proposed the MetaNeXt structure which significantly improves the performance of the convolutional neural network in the classification task. 
Inspired by this idea, we combine rectangular self-calibration attention with the MetaNeXt structure to further improve feature representation. Specifically, batch normalization and MLP are added after the rectangular self-calibration attention to refine features. 
Finally, the residual connection is employed to further enhance feature reuse. The architecture of the RCM is shown in \figref{fig:network}. It also can be described as:
\begin{equation}
\rm \mathbf{F}_{out}=\rho(\xi_F(x, \xi_C(H_P(x)\oplus V_P(x))))  + x,
\end{equation}
where $\oplus$ indicates broadcast addition. $H_P$ and $V_P$ represents Horizontal Pooling and Vertical Pooling. $\rho$ refers to BN and MLP.

\begin{figure}[t]
	\centering
	\includegraphics[width=0.8\columnwidth]{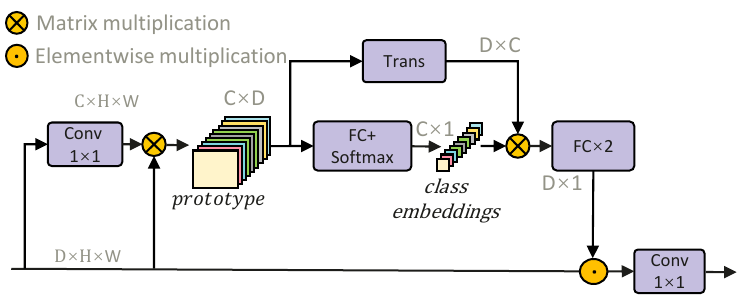}
	\caption{Dynamic Prototype Guided Head.}
	\label{DPGHead}
\end{figure}

\subsection{Dynamic Prototype Guided Head}
To improve the classification of foreground objects, the Dynamic Prototype Guided head is proposed, which generates the dynamic prototype to embed class information. The class embedding feature is projected into the pixel feature space, weighting the pixel features to enhance the distinction between different classes. The overall architecture of DPG Head is shown in Fig.~\ref{DPGHead}.

First, we project features into the class feature space. Then, the feature in class space and the feature in pixel space are multiplied to obtain the dynamic prototype. The dynamic prototype can reflect the feature distribution of different classes on each image, and it follows the dynamic change of the input. The process of generating the dynamic prototype can be expressed as:
\begin{equation}
\rm \mathbf{F}_p= \delta_{D\rightarrow C}(\mathbf{F}_x) \otimes \mathbf{F}_x, 
\end{equation}
where $\otimes$ is the matrix multiplication operation. $\rm \mathbf{F}_p$ represents the prototype.

Then, the DPG head further embeds class information to enhance the distinction between different classes. A fully connected layer is applied to compress the prototype to C$\times$1 dimension, effectively embedding the class information and yielding a class embedding vector. Softmax is applied to constrain the value between 0 and 1. The class embedding vector C$\times$1 represents the global information in each class. 
To project the class embedding vector into the pixel feature space, we multiply the class embedding vector by the transposed prototype. With the class embedding enhancement, the new attention vector has a stronger ability to distinguish the classes.
Then, two fully connected layers are used to capture the context in the attention vector. Layer Normalization is utilized to normalize features between two layers.
The calculation process of the class embedding vector is summarized by:
\begin{equation}
 \rm \mathbf{F}_{gp}=  Softmax\left(\delta_{D\rightarrow 1} \left(\mathbf{F}_p\right)\right) ,
\end{equation}
where $\delta$ represents the fully connected layer.

Finally, the attention vector is weighted to pixel features to emphasize important features and enhance feature representation. The calculation of the whole process is:
\begin{equation}
 \rm \mathbf{F}_{o}=  \delta \left(ReLU\left(LN\left(\delta \left(\mathbf{F}_p \otimes \mathbf{F}_{gp}\right)\right)\right)\right)\odot  \mathbf{F}_x,
\end{equation}
where $\rm LN $ and $\rm ReLU$ are the Layer Normalization and ReLU activation function. $\delta$ represents the fully connected layer. $\odot$ represents the broadcast Hadamard product. 
The calculated feature $\mathbf{F}_{o}$ is then fed into the classification convolution.

It is worth noting that the DPG Head is very lightweight. The prototype is compressed to C$\times$1 and the broadcast Hadamard product is used to spread attention to the global feature. These operations make the computational cost of the DPG Head very low. Current popular segmentation heads, \eg, OCRHead~\cite{ocrhead}, are in high computation complexity. Compared with them, the proposed DPG Head is more suitable for efficient semantic segmentation.

\section{Experiments}
\subsection{Datasets}
We conduct comprehensive experiments on the ADE20K~\cite{ade20k}, COCO-Stuff~\cite{coco-stuff}, and Pascal Context~\cite{pascal_context} datasets. The ADE20K~\cite{ade20k} is a challenging segmentation dataset. It contains more than 20,000 images from diverse environments and situations, covering 150 categories. 
COCO-Stuff~\cite{coco-stuff} is built from the COCO dataset by augmenting 164K images with pixel-level stuff annotation. It contains $172$ categories with $80$ things, $91$ stuff, and $1$ unlabeled class. The PASCAL Context~\cite{pascal_context} dataset is an extension of the PASCAL VOC 2010 detection challenge, which contains 59 classes. The dataset contains 4996 images in the training set and 5104 images in the test set.
\begin{table*}[tbp]
  \setlength{\tabcolsep}{1mm}
    \centering
    \caption{Results on ADE20K. We report the performance in terms of the mIoU, FLOPs, Parameter, and Throughput. 
    }
      \begin{tabular}{c|c|c|c|c}
      \toprule
      \textbf{Method} & \textbf{mIoU} & \textbf{FLOPs(G)} & \textbf{Param(M)} & \textbf{Thp.(Img/s)} \\
      \midrule
      DeeplabV3+\textcolor{red}{(ECCV'18)}~\cite{deeplabv3plus} & 34.0  & 69.4  & 15.4  & 63.0  \\
      Segformer-B0\textcolor{red}{(NeurIPS'21)}~\cite{segformer} & 37.4  & 8.4   & 3.8   & 117.1  \\
      FeedFormer-B0\textcolor{red}{(AAAI'23)}~\cite{feedformer} & 39.2  & 7.8   & 4.5   & 110.3 \\
      SegNeXt-T\textcolor{red}{(NeurIPS'22)}~\cite{segnext} & 41.1  & 6.6   & 4.3   & 123.5  \\
      Seaformer-L\textcolor{red}{(ICLR'23)}~\cite{seaformer} & 42.7  & 6.5   & 14.0  & 142.3 \\%147.0
      PEM-STDC1\textcolor{red}{(CVPR'24)}~\cite{pem} & 39.6  & 16.0   & 17.0  & -  \\
      \rowcolor[rgb]{ .894,  .875,  .925} CGRSeg-T (Ours) & 43.6  & 4.0   & 9.4   & 138.4  \\
      \midrule
      DeeplabV3+\textcolor{red}{(ECCV'18)}~\cite{deeplabv3plus} & 44.1  & 255.1  & 62.7  & 21.6  \\
      EncNet\textcolor{red}{(CVPR'18)}~\cite{encnet} & 44.7  & 218.8  & 68.6  & 23.4  \\
      CCNet\textcolor{red}{(ICCV'19)}~\cite{ccnet} & 45.2  & 278.4  & 68.9  & 23.2  \\
      Segformer-B1\textcolor{red}{(NeurIPS'21)}~\cite{segformer} & 42.2  & 15.9  & 13.7  & 96.0  \\
      SegNeXt-S\textcolor{red}{(NeurIPS'22)}~\cite{segnext} & 44.3  & 15.9  & 13.9  & 91.1  \\
      FeedFormer-B1\textcolor{red}{(AAAI'23)}~\cite{feedformer} & 41.0  & 10.0  & 4.6   & 87.2 \\
      PEM-STDC2\textcolor{red}{(CVPR'24)}~\cite{pem} & 45.0  & 19.3   & 21.0  & -  \\
      \rowcolor[rgb]{ .894,  .875,  .925} CGRSeg-B (Ours) & 45.5  & 7.6   & 18.1  & 98.4  \\
      \midrule
      Segformer-B2\textcolor{red}{(NeurIPS'21)}~\cite{segformer} & 46.5  & 62.4  & 27.5  & 70.4  \\
      MaskFormer\textcolor{red}{(NeurIPS'21)}~\cite{maskformer} & 46.7  & 55.0  & 42.0  & - \\
      Mask2Former\textcolor{red}{(CVPR'22)}~\cite{mask2former} & 47.7  & 74.0  & 47.0  & - \\
      FeedFormer-B2\textcolor{red}{(AAAI'23)}~\cite{feedformer} & 48.0  & 42.7  & 29.1  & 56.9 \\
      LRFormer-T\textcolor{red}{(arXiv'23)}~\cite{lrformer} &46.7 &17.0 & 13.0 & - \\
      \rowcolor[rgb]{ .894,  .875,  .925} CGRSeg-L (Ours) & 48.3  & 14.9  & 35.7  & 73.0  \\
      \bottomrule
      \end{tabular}%
    \label{tab:compare_sota}%
\end{table*}%

\subsection{Implementation details}
Our implementation is based on the public codebase mmsegmentation~\cite{mmsegmentation}. We use EfficientFormerV2~\cite{efficientformerv2} as the backbone network. For the ADE20K dataset, we follow SegNeXt~\cite{segnext} to use 160K scheduler and the batch size is 16. The initial learning rate is 0.00012 and the weight decay is 0.01. A “poly” learning rate scheduled with factor 1.0 is adopted. The training images are resized to 1024$\times$512 for ADE20K. For the COCO-Stuff dataset, we use 80K scheduler and the batch size is 8. The initial learning rate is 0.00012 and the weight decay is 0.01. The training images are resized to 1024$\times$512. For the Pascal Context dataset, the training iteration is also 80K with batch size 16. The initial learning rate is 0.00006 and the weight decay is 0.01. The training images are resized to 480$\times$480. For all three datasets, we use the default data augmentation strategy in mmsegmentation. The throughput is evaluated on an NVIDIA Tesla V100 GPU with the maximum batch size.

\subsection{Comparisons with the State-of-the-art Methods}
\begin{table}[tbp]
  \setlength{\tabcolsep}{2mm}
  \centering
  \caption{Performance comparison of state-of-the-art methods on COCO-Stuff dataset.}
    \begin{tabular}{c|c|c|c}
    \toprule
    \textbf{Method} & \textbf{mIoU($\%$)} & \textbf{FLOPs(G)} & \textbf{Param(M)} \\
    \midrule
    SegFormer-B0~\cite{segformer} & 35.6  & 8.4   & 3.8  \\
    SegNeXt-T~\cite{segnext}  & 38.7  & 6.6   & 4.3  \\
    \rowcolor[rgb]{ .894,  .875,  .925} CGRSeg-T  & 42.2  & 4.0   & 9.4  \\
    \midrule
    HRFormer-S~\cite{hrformer} & 37.9  & 109.5  & 13.5  \\
    SegFormer-B1~\cite{segformer} & 40.2  & 15.9  & 13.7  \\
    SegNeXt-S~\cite{segnext} & 42.2  & 15.9  & 13.9  \\
    \rowcolor[rgb]{ .894,  .875,  .925} CGRSeg-B  & 43.5  & 7.6   & 18.1  \\
    \midrule
    HRFormer-B~\cite{hrformer} & 42.4  & 280.0  & 56.2  \\
    LRFormer-T~\cite{lrformer} &43.9 & 17.0 &13.0\\
    SegFormer-B2~\cite{segformer} & 44.6  & 62.4  & 27.5  \\
    SegNeXt-B~\cite{segnext} & 45.8  & 34.9  & 27.6 \\
    \rowcolor[rgb]{ .894,  .875,  .925} CGRSeg-L & 46.0  & 14.9  & 35.7  \\
    \bottomrule
    \end{tabular}%
  \label{tab:cocostuff}%
\end{table}%
\textbf{Results on ADE20K.} As shown in \tabref{tab:compare_sota}, we compare the proposed method with various state-of-the-art models on the ADE20K dataset.
We report the performance of our model with three variant sizes, \ie, the tiny model, the base model, and the large model.

For the tiny models, the proposed CGRSeg-T achieves $43.6\%$ mIoU with only $4.0$ GFLOPs.
It performs better than all other CNN-based models and transformer-based models in terms of the mIoU and FLOPs.
For example, it exceeds the previous best model SeaFormer-L by $+0.9\%$ mIoU with $2.5$ GFLOPs lower computation cost and $4.6$M lower parameters.
As for the base models, the proposed CGRSeg-B yields $45.5\%$ mIoU with $7.6$ GFLOPs.
Note the computation cost of our model is only 39\% of the best compared PEM-STDC2 ($19.3$ GFLOPs) model, but the performance is $+0.5\%$ mIoU better.
Among the large models, our CGRSeg-L model ($15.1$ GFLOPs) outperforms MaskFormer (Swin-T) and Mask2Former (Swin-T) over $+1.6\%$/$+0.6\%$ mIoU, but only with around $1/4$ computational cost, which demonstrates the high efficiency of our model.

Based on the above experimental results, we can conclude that the proposed \ourmodel~can achieve higher performance with less computation cost. 
This is because the proposed lightweight RCM can make the model focus on the foreground objects, improving the feature representation. Additionally, the DPG Head also improves the classification of the foreground objects.

\textbf{Results on COCO-Stuff.} We compare the proposed \ourmodel~with previous models on COCO-Stuff~\cite{cocostuff} dataset in \tabref{tab:cocostuff}.
The performance of our model can exceed the previous best methods SegNeXt~\cite{segnext} with less computation cost.
For example, the tiny (T), base (B), and large (L) versions of \ourmodel~exceed the best-compared method by at least $+3.5\%$, $+1.3\%$ and $+0.2\%$ mIoU with only $61\%$, $48\%$, and $43\%$ FLOPs. Additionally, CGRSeg-L surpasses LRFormer-T by 2.1$\%$ in mIoU and outperforms it by 2.1 G in terms of FLOPs.

\textbf{Results on Pascal Context.} We compare the proposed \ourmodel~with state-of-the-art models on the Pascal Context dataset. As shown in \tabref{pascal_context}, our CGRSeg achieves better performance than all other compared methods.
For example, our \ourmodel-T model obtains $54.1\%$ mIOU with only $4.0$ GFLOPs, which is $+0.8\%$ better than SegNeXt-T~\cite{segnext} with $6.6$ FLOPs. When compared with the larger model SegNeXt-S, our model still achieves $+0.4\%$ better mIoU with about only $48\%$ FLOPs. Besides, \ourmodel-B also exceeds Senformer~\cite{senformer} $2.2\%$ mIoU. 

\begin{table}[tbp]
  \setlength{\tabcolsep}{1mm}
  \centering
  \caption{Performance comparison of state-of-the-art methods on Pascal Context dataset. The FLOPs is calculated with the input size of 512$\times$512. MS stands for multi-scale test setup.} 
    \begin{tabular}{c|c|c|c}
    \toprule
    \textbf{Method} & \textbf{mIoU(MS,\%)} & \textbf{FLOPs(G)} & \textbf{Param(M)} \\
    \midrule
    DANet~\cite{danet} & 52.6  & 277.7  & 69.1  \\
    EncNet~\cite{encnet} & 52.6  & -     & - \\
    EMANet~\cite{emanet} & 53.1  & 246.1  & 53.1  \\
    SegNeXt-T~\cite{segnext} & 53.3  & 6.6   & 4.3  \\
    \rowcolor[rgb]{ .894,  .875,  .925} CGRSeg-T & 54.1  & 4.0   & 9.4  \\
    \midrule
    HamNet~\cite{hamburgerhead}& 55.2  & 277.9  & 69.1  \\
    HRNet(OCR)~\cite{hrnet} & 56.2  & -     & 74.5  \\
    HRFormer-S~\cite{hrformer} & 54.6  & -     & - \\
    Senformer~\cite{senformer} & 54.3  & 179.0  & 55.0  \\
    SegNeXt-S~\cite{segnext} & 56.1  & 15.9  & 13.9  \\
    \rowcolor[rgb]{ .894,  .875,  .925} CGRSeg-B & 56.5  & 7.6   & 18.1  \\
    \midrule
    Senformer~\cite{senformer} & 56.6  & 199.0  & 79.0  \\
    \rowcolor[rgb]{ .894,  .875,  .925} CGRSeg-L & 58.5  & 14.9  & 35.7  \\
    \bottomrule
    \end{tabular}%
  \label{pascal_context}%
\end{table}%

\begin{table*}[tbp]
  \setlength{\tabcolsep}{1mm}
    \centering
    \caption{The ablation analysis of the RCM and the DPG Head. 
    Spatial feature reconstruction and pyramid context extraction are denoted as SFR and PCE, respectively.
    }
      \begin{tabular}{c|c|c|c|c|c}
      \toprule
      \textbf{RCM(SFR)} & \textbf{RCM(PCE)} & \textbf{DPG Head} & \textbf{mIoU(\%)} & \textbf{FLOPs(G)} & \textbf{Param(M)} \\
      \midrule
      $\times$   &  $\times$   &  $\times$   & 40.86  & 3.56  & 6.08  \\
      $\times$  & $\times$    & \checkmark    & 41.34  & 3.64  & 6.08  \\
      $\times$      & \checkmark    & \checkmark     & 42.57  & 3.83  & 9.08  \\
      \checkmark & \checkmark     &    $\times$   & 42.56  & 3.92  & 9.36  \\
      \checkmark     & \checkmark    & \checkmark     & 43.60  & 4.00  & 9.36  \\
      \bottomrule
      \end{tabular}%
    \label{tab:main_ablation}%
  \end{table*}%

\subsection{Ablation Study}
\myPara{The ablation analysis of the proposed modules.}
The two core components of the proposed \ourmodel~are the RCM and the final DPG head.
To validate the effectiveness of the two components, in \tabref{tab:main_ablation}, we make an ablation study of the RCM and DPG head.
The base model is the first row of the table without any elaborate designs.
The proposed RCM is used for pyramid context extraction (PCE) and spatial feature reconstruction (SFR).
In the base model, we replace the RCM for pyramid context extraction by $1\times 1$ convolutions and directly drop RCMs for feature reconstruction.
DPG Head is directly dropped.
As shown in the first row of the table, when not using the two modules, the performance is only $40.86\%$ mIoU.
In the third row, adding the RCM for pyramid context extraction can increase the performance by $1.23\%$ mIoU with only an extra $0.19$ GFLOPs.
Adding the RCM for feature reconstruction can increase the performance by $1.13\%$ mIoU with only an extra $0.17$ GFLOPs.
Adding the DPG Head, \ie, the second row in the table, the performance will reach $41.34\%$ from $40.86\%$.
When combined with the two modules, the final performance can achieve $43.60\%$ mIoU, achieving a $2.74\%$ improvement over the base model.

\begin{table}[tbp]
  \begin{minipage}[t]{0.45\textwidth}
    \caption{Performance comparison with other advanced segmentation heads. The DPG head significantly outperforms other heads. }
    \centering
    \setlength{\tabcolsep}{1.5mm}
    \scalebox{0.9}{
    \begin{tabular}{c|c|c}
    \toprule
    \textbf{Head} & \textbf{mIoU($\%$)} & \textbf{FLOPs(G)}  \\
    \midrule
    None-local & 40.9  & 4.05   \\
    DAHead & 42.6  & 4.05    \\
    HamHead & 42.1  & 4.05   \\
    GC block & 42.7  & 3.92   \\
    SE block & 42.6  & 3.92  \\
    \rowcolor[rgb]{ .894,  .875,  .925} DPG Head & 43.6  & 4.00  \\
    \bottomrule
    \end{tabular}%
    }
    \label{cmp_head}%
    \end{minipage}
    \begin{minipage}[t]{0.55\textwidth}
      \caption{The performance comparison of RCA and other advanced attention block. RCA outperforms other attention modules without increasing Flops.}
      \centering
      \scalebox{0.9}{
      \begin{tabular}{c|c|c|c}
      \toprule
      \textbf{Method} & \textbf{mIoU($\%$)} & \textbf{FLOPs(G)} & \textbf{Param(M)} \\
      \midrule
      Self-Att & 39.9  & 4.1   & 9.6  \\
      ConvNext & 41.6  & 4.0   & 9.4  \\
      InceptionNext & 41.6  & 4.0   & 9.3  \\
      CoordAtt & 41.5  & 4.0   & 9.9  \\
      GatherExcite & 41.7  & 4.0   & 11.0  \\
      \rowcolor[rgb]{ .894,  .875,  .925} RCA(Ours) & 43.6  & 4.0   & 9.4  \\
      \bottomrule
      \end{tabular}%
      }
    \label{cmp_sca}%
    \end{minipage}
  \end{table}

\par
\myPara{Comparison with other advanced heads.} It is well known that a lightweight segmentation head is important for building an efficient segmentation framework.
Thus, we elaborately design the DPG head to meet the low FLOPs and fast speed needs.
As shown in \tabref{cmp_head}, the proposed DPG Head is compared with other popular and alternative segmentation heads.
To get the performance of other heads, we directly replace the DPG Head with them in the proposed CGRSeg-T. For GCHead, it is with the least computation cost, but the mIoU of it is poorer than our head. Compared with HamHead, our head exceeds it by 1.5$\%$ mIoU, but FLOPs drop 0.05 G.
Due to the proposed efficient DPG Head, the total computation cost of the proposed CGRSeg-T in \tabref{tab:compare_sota} is with only $4.0$ GFLOPs but can achieve the best mIoU, compared with other state-of-the-art methods. The proposed DPG Head is efficient because of the dynamic prototype guidance.

\myPara{Comparisons of alternative blocks of the RCA.} As shown in \tabref{cmp_sca}, we compare the performance of other alternative sub-blocks of the RCA, \eg, self-attention~\cite{topformer}, InceptionNext~\cite{inceptionnext}, ConvNext~\cite{convnext}, CoordAtt Block~\cite{coordatt} and GatherExcite Block~\cite{ge}.
To compare them, we directly replace the RCM with other blocks.
It can be found that the proposed CGRSeg performs better than the CoordAtt Block and GatherExcite Block by increasing $2.1\%$ and $1.9\%$ mIoU.

\myPara{Comparisons with other backbone.} To further validate the sophistication of our method, we replace the backbone with MSCAN which is the backbone of SegNeXt. As shown in Table~\ref{backbone}, \ourmodel~still increases mIoU by 1.5$\%$ and FLOPs drop by 0.7G compared to SegNeXt. This shows that our CGRSeg framework is efficient and can be adapted to different backbone networks.

\begin{table}[tbp]
  \setlength{\tabcolsep}{1.5mm}
  \centering
  \caption{Performance comparison with other backbone settings.}
    \begin{tabular}{c|c|c|c|c}
    \toprule
    \textbf{Method} & \textbf{Backbone} & \textbf{mIoU($\%$)} & \textbf{FLOPs(G)} & \textbf{Param(M)} \\
    \midrule
    SegNeXt & MSCAN-T & 41.1  & 6.6   & 4.3  \\
    CGRSeg-T & MSCAN-T & 42.6  & 5.9   & 9.7  \\
    \bottomrule
    \end{tabular}%
  \label{backbone}%
\end{table}%

\begin{table}[tbp]
  \setlength{\tabcolsep}{1.5mm}
  \centering
  \caption{The effect of Broadcast Addition and Broadcast Multiplication on Rectangular Self-Calibration Attention Attention.}
    \begin{tabular}{c|c|c|c}
    \toprule
    \textbf{Key Area Modeling} & \textbf{mIoU($\%$)} & \textbf{FLOPs(G)} & \textbf{Param(M)} \\
    \midrule
    None    & 42.22 & 4.00  & 9.4  \\
    Mul   & 43.08 & 4.00  & 9.4  \\
    Add   & 43.61 & 4.00  & 9.4  \\
    \bottomrule
    \end{tabular}%
  \label{cmp_add}%
\end{table}%

\begin{table}[tbp]
  \begin{minipage}[t]{0.5\textwidth}
    \caption{Convolution kernel selection in shape self-calibration function.}
    \centering
    \begin{tabular}{c|c}
      \toprule
      \textbf{Kernel Size} & \textbf{mIoU($\%$)} \\
      \midrule
      None  & 41.27 \\
      $1\times 5 $, $5\times 1$ & 42.51 \\
      $1\times 7 $, $7\times 1$ & 43.39 \\
      $1\times 9 $, $9\times 1$ & 43.50 \\
      $1\times 11$ , $11\times$ 1    & 43.61 \\
    \bottomrule
    \end{tabular}%
    \label{kernel_size}
    \end{minipage}
    \begin{minipage}[t]{0.5\textwidth}
    \caption{Convolution kernel selection for fusion functions in Rectangular Self-Calibration Attention Attention.}
    \centering
    \begin{tabular}{c|c}
      \toprule
      \textbf{Local Conv} & \textbf{mIoU($\%$)} \\
      \midrule
      None  & 42.40  \\
      $1\times 1 $   & 41.70 \\
      $3\times 3 $   & 43.39 \\
      $5 \times 5 $   & 42.99 \\
      \bottomrule
      \end{tabular}%
    \label{local_size}
    \end{minipage}
  \end{table}
\myPara{Comparison of addition and multiplication in RCA.} As shown in \tabref{cmp_add}, we compare the performance of Addition and Multiplication on RCA. It can be found that explicit modeling of key regions can effectively improve segmentation performance. Besides, the performance of addition is better than that of multiplication by 0.53$\%$ mIoU. 

\myPara{The effectiveness of the kernel size in shape self-calibration
function.}
To investigate the effectiveness of the kernel size in the shape self-calibration function, we conduct a series of experiments in the \tabref{kernel_size}.
As the size of the convolution kernel increases, the segmentation precision also increases. When the kernel sizes are $1\times 11$, and $11\times 1$, the performance is best. 
Because increasing the kernel size further leads to the increase of FLOPs, we do not increase the convolution kernel.
Finally, we choose the groups of $1\times 11$, and $11\times 1$ kernel sizes in our model.

\myPara{Convolution kernel selection for fusion functions in RCA.}
To investigate the effectiveness of the kernel size in the fusion functions, we conduct a series of experiments in the \tabref{local_size}.
As shown in the table, when the kernel size is $3\times 3$, the performance is best. 
Thus, we finally choose $3\times 3$ kernel size in the fusion function.

\begin{figure}[t]
  \centering
  \includegraphics[width=0.96\textwidth]{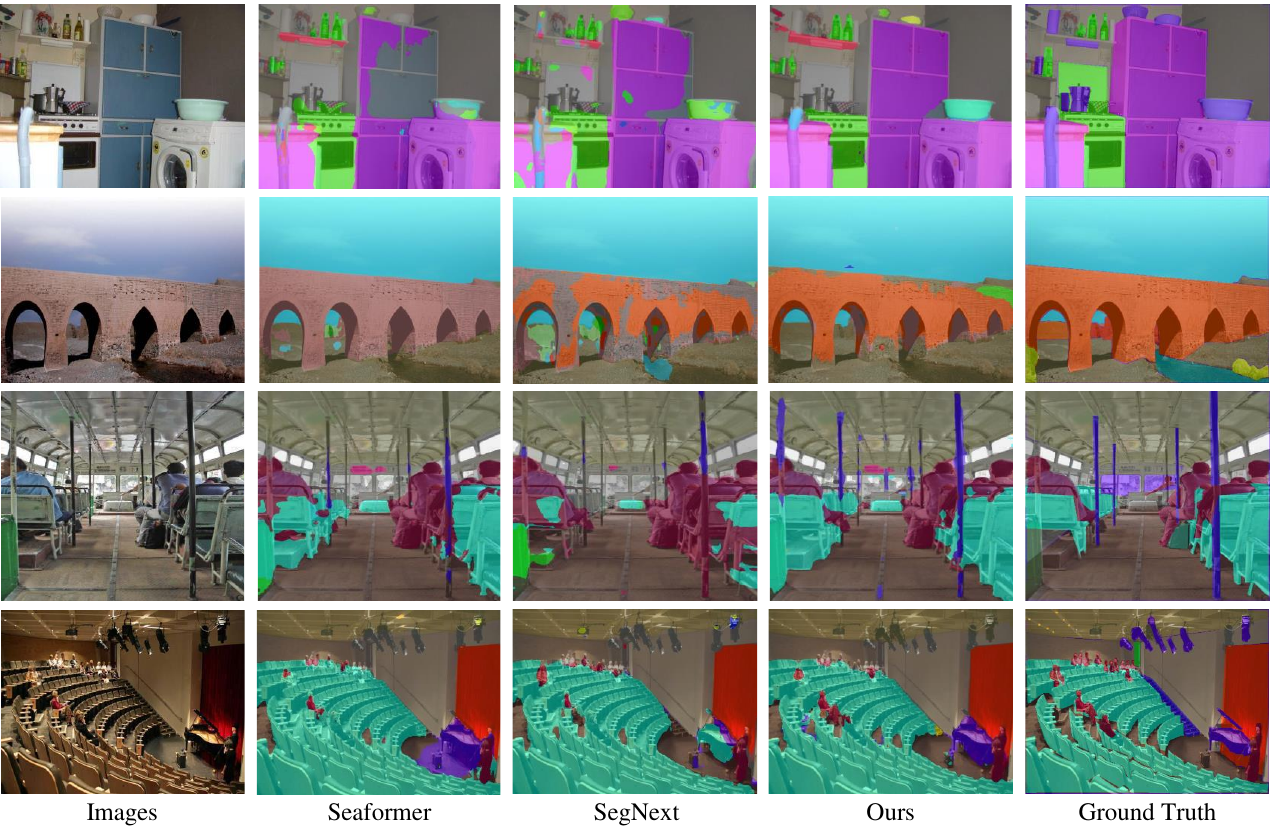}
  \caption{Qualitative Comparison of CGRSeg-T on the ADE20K dataset.}
  \label{output}
\end{figure}

\myPara{Visualization.}
The prediction results of SegNeXt and CGRSeg-T are visualized in Fig.~\ref{output}. 
The mask of CGRSeg is closer to the ground truth than that of other methods.
It can be found that our CGRSeg-T achieves more complete segmentation masks with fewer errors.

\section{Conclusion}
In this paper, we present CGRSeg, an efficient yet competitive segmentation framework consisting of pyramid context extraction and spatial feature reconstruction. To implement these two parts, an RCM is carefully designed to make the model focus on the foreground and capture the axial global context. Furthermore, the DPG Head is proposed to significantly improve the feature representation with only a small amount of computation. Our \ourmodel~is extensively evaluated on ADE20K, COCO-Stuff, and Pascal Context benchmarks, achieving state-of-the-art segmentation performance on all three datasets. 
\bibliographystyle{splncs04}
\bibliography{ref}

\begin{thebibliography}{10}
\providecommand{\url}[1]{\texttt{#1}}
\providecommand{\urlprefix}{URL }
\providecommand{\doi}[1]{https://doi.org/#1}

\bibitem{senformer}
Bousselham, W., Thibault, G., Pagano, L., Machireddy, A., Gray, J., Chang,
  Y.H., Song, X.: Efficient self-ensemble for semantic segmentation. British
  Machine Vision Conference  (2022)

\bibitem{coco-stuff}
Caesar, H., Uijlings, J., Ferrari, V.: Coco-stuff: Thing and stuff classes in
  context. In: Proceedings of the IEEE conference on computer vision and
  pattern recognition. pp. 1209--1218 (2018)

\bibitem{cocostuff}
Caesar, H., Uijlings, J., Ferrari, V.: Coco-stuff: Thing and stuff classes in
  context. In: CVPR. pp. 1209--1218 (2018)

\bibitem{gcnet}
Cao, Y., Xu, J., Lin, S., Wei, F., Hu, H.: Gcnet: Non-local networks meet
  squeeze-excitation networks and beyond. In: ICCV Workshops. pp.~0--0 (2019)

\bibitem{pem}
Cavagnero, N., Rosi, G., Ruttano, C., Pistilli, F., Ciccone, M., Averta, G.,
  Cermelli, F.: Pem: Prototype-based efficient maskformer for image
  segmentation. In: Proceedings of the IEEE/CVF conference on computer vision
  and pattern recognition (2024)

\bibitem{deeplab}
Chen, L.C., Papandreou, G., Kokkinos, I., Murphy, K., Yuille, A.L.: Deeplab:
  Semantic image segmentation with deep convolutional nets, atrous convolution,
  and fully connected crfs. IEEE T-PAMI  \textbf{40}(4),  834--848 (2017)

\bibitem{deeplabv3}
Chen, L.C., Papandreou, G., Schroff, F., Adam, H.: Rethinking atrous
  convolution for semantic image segmentation. arXiv preprint arXiv:1706.05587
  (2017)

\bibitem{deeplabv2}
Chen, L.C., Zhu, Y., Papandreou, G., Schroff, F., Adam, H.: Encoder-decoder
  with atrous separable convolution for semantic image segmentation. In: ECCV.
  pp. 801--818 (2018)

\bibitem{deeplabv3plus}
Chen, L.C., Zhu, Y., Papandreou, G., Schroff, F., Adam, H.: Encoder-decoder
  with atrous separable convolution for semantic image segmentation. In:
  Proceedings of the European conference on computer vision (ECCV). pp.
  801--818 (2018)

\bibitem{mask2former}
Cheng, B., Misra, I., Schwing, A.G., Kirillov, A., Girdhar, R.:
  Masked-attention mask transformer for universal image segmentation. In:
  Proceedings of the IEEE/CVF conference on computer vision and pattern
  recognition. pp. 1290--1299 (2022)

\bibitem{maskformer}
Cheng, B., Schwing, A., Kirillov, A.: Per-pixel classification is not all you
  need for semantic segmentation. NeurIPS  \textbf{34},  17864--17875 (2021)

\bibitem{mmsegmentation}
Contributors, M.: {MMSegmentation}: Openmmlab semantic segmentation toolbox and
  benchmark. \url{https://github.com/open-mmlab/mmsegmentation} (2020)

\bibitem{stdnc}
Fan, M., Lai, S., Huang, J., Wei, X., Chai, Z., Luo, J., Wei, X.: Rethinking
  bisenet for real-time semantic segmentation. In: CVPR. pp. 9716--9725 (2021)

\bibitem{danet}
Fu, J., Liu, J., Tian, H., Li, Y., Bao, Y., Fang, Z., Lu, H.: Dual attention
  network for scene segmentation. In: CVPR. pp. 3146--3154 (2019)

\bibitem{hamburgerhead}
Geng, Z., Guo, M.H., Chen, H., Li, X., Wei, K., Lin, Z.: Is attention better
  than matrix decomposition? arXiv preprint arXiv:2109.04553  (2021)

\bibitem{segnext}
Guo, M.H., Lu, C.Z., Hou, Q., Liu, Z.N., Cheng, M.M., Hu, S.m.: Segnext:
  Rethinking convolutional attention design for semantic segmentation. In:
  NeurIPS (2022)

\bibitem{coordatt}
Hou, Q., Zhou, D., Feng, J.: Coordinate attention for efficient mobile network
  design. In: Proceedings of the IEEE/CVF conference on computer vision and
  pattern recognition. pp. 13713--13722 (2021)

\bibitem{ge}
Hu, J., Shen, L., Albanie, S., Sun, G., Vedaldi, A.: Gather-excite: Exploiting
  feature context in convolutional neural networks. Advances in neural
  information processing systems  \textbf{31} (2018)

\bibitem{ccnet}
Huang, Z., Wang, X., Huang, L., Huang, C., Wei, Y., Liu, W.: Ccnet: Criss-cross
  attention for semantic segmentation. In: ICCV. pp. 603--612 (2019)

\bibitem{dfanet}
Li, H., Xiong, P., Fan, H., Sun, J.: Dfanet: Deep feature aggregation for
  real-time semantic segmentation. In: CVPR. pp. 9522--9531 (2019)

\bibitem{hssn}
Li, L., Wang, W., Zhou, T., Quan, R., Yang, Y.: Semantic hierarchy-aware
  segmentation. IEEE Transactions on Pattern Analysis and Machine Intelligence
  (2023)

\bibitem{emanet}
Li, X., Zhong, Z., Wu, J., Yang, Y., Lin, Z., Liu, H.: Expectation-maximization
  attention networks for semantic segmentation. In: Proceedings of the IEEE/CVF
  International Conference on Computer Vision. pp. 9167--9176 (2019)

\bibitem{efficientformerv2}
Li, Y., Hu, J., Wen, Y., Evangelidis, G., Salahi, K., Wang, Y., Tulyakov, S.,
  Ren, J.: Rethinking vision transformers for mobilenet size and speed. arXiv
  preprint arXiv:2212.08059  (2022)

\bibitem{clustseg}
Liang, J., Zhou, T., Liu, D., Wang, W.: Clustseg: Clustering for universal
  segmentation. arXiv preprint arXiv:2305.02187  (2023)

\bibitem{autodeeplab}
Liu, C., Chen, L.C., Schroff, F., Adam, H., Hua, W., Yuille, A.L., Fei-Fei, L.:
  Auto-deeplab: Hierarchical neural architecture search for semantic image
  segmentation. In: CVPR. pp. 82--92 (2019)

\bibitem{convnext}
Liu, Z., Mao, H., Wu, C.Y., Feichtenhofer, C., Darrell, T., Xie, S.: A convnet
  for the 2020s. In: CVPR. pp. 11976--11986 (2022)

\bibitem{fcn}
Long, J., Shelhamer, E., Darrell, T.: Fully convolutional networks for semantic
  segmentation. In: CVPR. pp. 3431--3440 (2015)

\bibitem{pascal_context}
Mottaghi, R., Chen, X., Liu, X., Cho, N.G., Lee, S.W., Fidler, S., Urtasun, R.,
  Yuille, A.: The role of context for object detection and semantic
  segmentation in the wild. In: Proceedings of the IEEE conference on computer
  vision and pattern recognition. pp. 891--898 (2014)

\bibitem{feedformer}
Shim, J.h., Yu, H., Kong, K., Kang, S.J.: Feedformer: Revisiting transformer
  decoder for efficient semantic segmentation. In: Proceedings of the AAAI
  Conference on Artificial Intelligence. vol.~37, pp. 2263--2271 (2023)

\bibitem{seaformer}
Wan, Q., Huang, Z., Lu, J., Yu, G., Zhang, L.: Seaformer: Squeeze-enhanced
  axial transformer for mobile semantic segmentation  (2023)

\bibitem{hrnet}
Wang, J., Sun, K., Cheng, T., Jiang, B., Deng, C., Zhao, Y., Liu, D., Mu, Y.,
  Tan, M., Wang, X., Liu, W., Xiao, B.: Deep high-resolution representation
  learning for visual recognition. TPAMI  (2019)

\bibitem{contrastseg}
Wang, W., Zhou, T., Yu, F., Dai, J., Konukoglu, E., Van~Gool, L.: Exploring
  cross-image pixel contrast for semantic segmentation. In: Proceedings of the
  IEEE/CVF international conference on computer vision. pp. 7303--7313 (2021)

\bibitem{nonlocal}
Wang, X., Girshick, R., Gupta, A., He, K.: Non-local neural networks. In: CVPR.
  pp. 7794--7803 (2018)

\bibitem{lrformer}
Wu, Y.H., Zhang, S.C., Liu, Y., Zhang, L., Zhan, X., Zhou, D., Feng, J., Cheng,
  M.M., Zhen, L.: Low-resolution self-attention for semantic segmentation.
  arXiv preprint arXiv:2310.05026  (2023)

\bibitem{segformer}
Xie, E., Wang, W., Yu, Z., Anandkumar, A., Alvarez, J.M., Luo, P.: Segformer:
  Simple and efficient design for semantic segmentation with transformers.
  NeurIPS  \textbf{34},  12077--12090 (2021)

\bibitem{bisenet}
Yu, C., Wang, J., Peng, C., Gao, C., Yu, G., Sang, N.: Bisenet: Bilateral
  segmentation network for real-time semantic segmentation. In: ECCV. pp.
  325--341 (2018)

\bibitem{inceptionnext}
Yu, W., Zhou, P., Yan, S., Wang, X.: Inceptionnext: When inception meets
  convnext. arXiv preprint arXiv:2303.16900  (2023)

\bibitem{ocrhead}
Yuan, Y., Chen, X., Wang, J.: Object-contextual representations for semantic
  segmentation. In: ECCV. pp. 173--190 (2020)

\bibitem{hrformer}
Yuan, Y., Fu, R., Huang, L., Lin, W., Zhang, C., Chen, X., Wang, J.: Hrformer:
  High-resolution vision transformer for dense predict. NeurIPS  \textbf{34},
  7281--7293 (2021)

\bibitem{segvit}
Zhang, B., Tian, Z., Tang, Q., Chu, X., Wei, X., Shen, C., et~al.: Segvit:
  Semantic segmentation with plain vision transformers pp. 4971--4982 (2022)

\bibitem{encnet}
Zhang, H., Dana, K., Shi, J., Zhang, Z., Wang, X., Tyagi, A., Agrawal, A.:
  Context encoding for semantic segmentation. In: CVPR. pp. 7151--7160 (2018)

\bibitem{topformer}
Zhang, W., Huang, Z., Luo, G., Chen, T., Wang, X., Liu, W., Yu, G., Shen, C.:
  Topformer: Token pyramid transformer for mobile semantic segmentation. In:
  CVPR. pp. 12083--12093 (2022)

\bibitem{pspnet}
Zhao, H., Shi, J., Qi, X., Wang, X., Jia, J.: Pyramid scene parsing network.
  In: CVPR. pp. 2881--2890 (2017)

\bibitem{setr}
Zheng, S., Lu, J., Zhao, H., Zhu, X., Luo, Z., Wang, Y., Fu, Y., Feng, J.,
  Xiang, T., Torr, P.H., et~al.: Rethinking semantic segmentation from a
  sequence-to-sequence perspective with transformers. In: CVPR. pp. 6881--6890
  (2021)

\bibitem{ade20k}
Zhou, B., Zhao, H., Puig, X., Fidler, S., Barriuso, A., Torralba, A.: Scene
  parsing through ade20k dataset. In: CVPR. pp. 633--641 (2017)

\bibitem{protoseg}
Zhou, T., Wang, W., Konukoglu, E., Van~Gool, L.: Rethinking semantic
  segmentation: A prototype view. In: Proceedings of the IEEE/CVF Conference on
  Computer Vision and Pattern Recognition. pp. 2582--2593 (2022)

\end{thebibliography}

\end{document}